\documentclass[runningheads]{llncs}
\usepackage[T1]{fontenc}
\usepackage{graphicx}
\usepackage{amssymb}
\usepackage{booktabs}
\usepackage{multirow}
\usepackage{multicol}
\usepackage{pifont}
\usepackage{marvosym}
\usepackage{subfigure}

\begin{document}
\title{CMFN: Cross-Modal Fusion Network for Irregular Scene Text Recognition}
%
%
\author{Jinzhi Zheng\inst{1,3}\orcidID{0000-0002-7954-8014} \and
Ruyi Ji \textsuperscript{\Letter} \inst{2}\orcidID{0000-0001-8918-0981} \and
Libo Zhang\inst{1,4}\orcidID{0000-0001-8450-0958} \and
Yanjun Wu\inst{1,4} \and
Chen Zhao\inst{1,4}}
\authorrunning{J. Zheng et al.}
%
\institute{Intelligent Software Research Center, Institute of Software Chinese Academy of Sciences, Beijing, 100190, China \and 
The Laboratory of Cognition and Decision Intelligence for Complex Systems, Institute of Automation, Chinese Academy of Sciences, Beijing 100190, China \email{jrylovezd@gmail.com}\and
University of Chinese Academy of Sciences, Beijing, China \and
State Key Laboratory of Computer Science, Institute of Software Chinese Academy of Sciences, Beijing, 100190, China  \email{\{jinzhi2018,Libo,yanjun,zhaochen\}@iscas.ac.cn}}
\maketitle              
\begin{abstract}
Scene text recognition, as a cross-modal task involving vision and text, is an important research topic in computer vision. Most existing methods use language models to extract semantic information for optimizing visual recognition. However, the guidance of visual cues is ignored in the process of semantic mining, which limits the performance of the algorithm in recognizing irregular scene text. To tackle this issue, we propose a novel cross-modal fusion network (CMFN) for irregular scene text recognition, which incorporates visual cues into the semantic mining process. Specifically, CMFN consists of a position self-enhanced encoder, a visual recognition branch and an iterative semantic recognition branch. The position self-enhanced encoder provides character sequence position encoding for both the visual recognition branch and the iterative semantic recognition branch. The visual recognition branch carries out visual recognition based on the visual features extracted by CNN and the position encoding information provided by the position self-enhanced encoder. The iterative semantic recognition branch, which consists of a language recognition module and a cross-modal fusion gate, simulates the way that human recognizes scene text and integrates cross-modal visual cues for text recognition. The experiments demonstrate that the proposed CMFN algorithm achieves comparable performance to state-of-the-art algorithms, indicating its effectiveness.

\keywords{Scene Text Recognition  \and Scene Text Understanding \and Neural Networks \and OCR}
\end{abstract}

\begin{figure}[htbp]
\centering
\includegraphics[width=10cm,height=5cm]{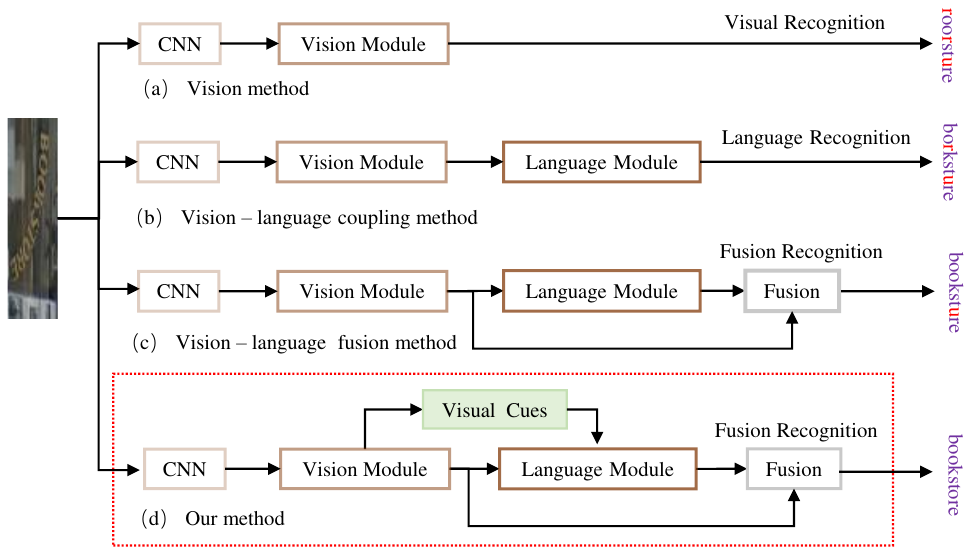}
\caption{Comparison of different scene text recognition methods related to our algorithm. (a) Visual recognition methods. (b) Recognition method of visual module series language module. (c) The method of visual module recognition is modified by the language module. (D) Our scene text recognition method(CMFN). Our CMFN fuses visual cues in the language module when mining semantic information.}
\label{Figure1.}
\end{figure}

\section{Introduction}
Scene text recognition, whose main task is to recognize text in image blocks \cite{2}, remains a research hotspot in the field of artificial intelligence because of its wide application scenarios \cite{47}, such as intelligent driving, visual question answering, image caption. This task still faces great challenges, mainly due to the complexity and diversity of scene text \cite{6}.

Most existing methods mainly consider scene text recognition as a sequence generation and a prediction task \cite{43,14}. In the early methods \cite{43,14,11}, CNN extracts visual features from the scene image. The vision module decodes the visual features and predicts the scene text (Fig. \ref{Figure1.}(a)). For example, TRBA \cite{43}, MORAN \cite{14}, SATRN \cite{19} and VisionLAN \cite{20} use LSTM or transformer module to decode the visual features extracted by CNN. This kind of vision method decodes the visual features but encodes less textual semantic information.

The text carries rich semantic information, which is of great significance in improving the accuracy of text recognition. To exploit the semantic information of text in the recognition process, several visual-language coupling methods \cite{44,21} have been proposed. This kind of method connects the language module after the vision module(Fig. \ref{Figure1.}(b)). For example, a language decoder set up in PIMNet \cite{21} extracts semantic information from previous predictive text for semantic recognition. The advantage is that the text semantic information can be mined in the recognition process, but the disadvantage is that the recognition result only depends on the performance of the language module, and visual recognition is only the intermediate result. In order to overcome this problem, some visual-language fusion methods \cite{9,2,6,45} have been gradually proposed(Fig. \ref{Figure1.} (c)).  After the language module, the fusion gate fuses the visual recognition of the vision module with the semantic recognition of the language module and outputs the fused recognition. The language module in these algorithms only excavates semantic information from the recognized text and ignores visual clues, which means that even if interesting visual clues are found in images or videos, the language module cannot use them to enhance its understanding of semantics.

When humans read scene text, if the visual features are not enough to recognize the text, they will extract semantic information based on the previous recognition. The process of extracting semantic information, not only relies on the context of previous recognition but also incorporates visual cues. It is important to note that the visual cues here are slightly different from explicit visual information such as color, shape, brightness, etc. They refer to more abstract signals or hints perceived from this explicit visual information. Inspired by this, this paper proposes a novel cross-modal fusion network(CMFN) to recognize irregular scene text. The contributions of this paper can be summarized as follows:
\begin{itemize}
\item
We propose a novel cross-modal fusion network that divides the recognition process into two stages: visual recognition and iterative semantic recognition, with cross-modal fusion in the iterative semantic recognition process.

\item  
We design a  position self-enhanced encoder to provide more efficient position coding information for the visual recognition branch and iterative semantic recognition branch.

\item 
In the iterative semantic recognition branch, we design a language module that fuses visual cues. It can alleviate the problem of over-reliance on visual recognition when language the module mining semantic information.

\item 
To verify the effectiveness of the proposed algorithm, abundant experiments are carried out on publicly available datasets. 
\end{itemize}

\begin{figure}[t]
\centering
\includegraphics[width=1\textwidth]{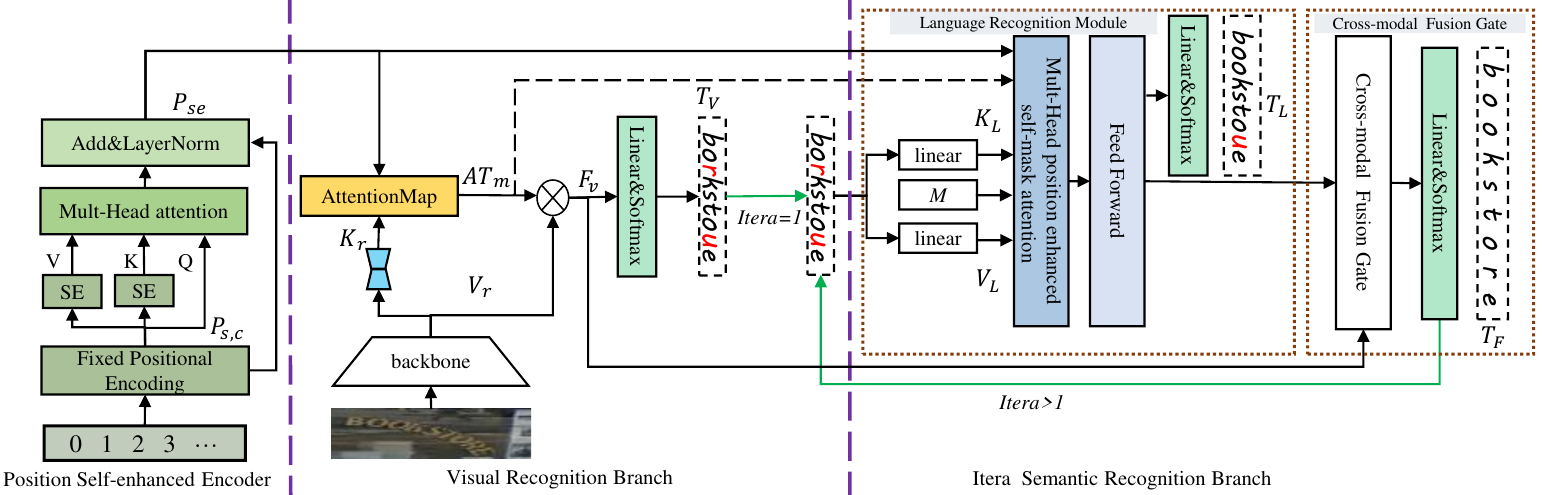}
\caption{The overall architecture of CMFN, comprises a position self-enhanced encoder, a visual recognition branch, and an iterative semantic recognition branch. 
The dashed arrow indicates the direction of attention maps $AT_m$ as visual cues transmission. The blue arrows represent the iterative process.}
\label{Figure2.}
\end{figure}

\section{Methodology}
\label{Overview}
The overall structure of CMFN is shown in Fig. \ref{Figure2.}. Given scene text image $I$ and pre-defined maximum text length $T$, the recognition process of scene text consists of three steps. Firstly, the position self-enhanced encoder encodes the character sequence information of the text and outputs position self-enhanced embeddings $P_{se}$. The character sequence information is an increasing sequence starting from 0 to $T$-1. Then, the visual branch extracts text visual features from the scene image $I$ and performs text visual recognition based on the visual features and the position self-enhanced embeddings $P_{se}$ and outputs the text prediction probability as text visual recognition $T_V$. In addition, attention map $AT_m$ is also output as visual cues. Finally, the iterative semantic recognition branch mines semantic information in an iterative way. The iterative semantic recognition branch consists of a language recognition module and a cross-modal fusion gate. The language recognition module integrates visual cues to mine semantic information and outputs the text language prediction probability as text language recognition $T_L$. The cross-modal fusion gate fuses visual and language features to output cross-modal text fusion prediction probability as text fusion recognition $T_F$. At the end of the last iteration, the text fusion recognition $T_F$ is output as the final recognition result.

\subsection{Position Self-enhanced Encoder}
\label{Position}
While the structure of learnable positional encoding \cite{48} is relatively simple, a large dataset is required for training. The fixed positional encoding \cite{23} uses constant to express position information, which can meet the requirements in some tasks with relatively low sensitivity to the position. However, in scene text recognition tasks, there are limited training samples and existing models are sensitive to character position information. We are inspired by \cite{9,23,6} to propose a position self-enhanced encoder whose structure is shown in Fig. \ref{Figure2.}.

The position self-enhanced encoder is designed to enhance the expression ability of character position information based on the correlation between the encoding feature dimensions. The following formula can be obtained:
\begin{equation}
E_\delta=\delta\left(CNN\left(pool\left(P_{s,c}\right)\right)\right)\ \ 
\end{equation}
\begin{equation}
P_e=P_{s,c}\ast\sigma\left(CNN\left(E_\delta\right)\right)+P_{s,c} 
\end{equation}
where $P_{s,c}\in \mathbb{R}^{T\times C_p} $ is fixed positional encoding, $C_p$ is the dimension of the position self-enhanced embedding, $pool$ is average pooling operation, $CNN$ is convolution operation, $\delta$ and $\sigma$ are relu and sigmoid activation functions, $P_e$ is the output of the self-enhance (SE) block. The formulaic expression of position self-enhanced encoder coding process:
\begin{equation}
M_p=softmax\left(\frac{QK^\mathsf{T}}{\sqrt{C_p}}\right)V+P_{s,c}
\end{equation}
\begin{equation}
P_{se}=LayNorm\left(M_p\right)\in \mathbb{R}^{T\times C_p} 
\end{equation}
where query $Q$ is from the fixed positional encoding $P_{s,c}$, key $K$ and value $V$ are from two self-enhanced block outputs $P_e$, respectively. $K^\mathsf{T}$ is the transpose of key $K$. $LayNorm$ represents layer normalization and $P_{se}$ is the output of position self-enhanced encoder. 
\subsection{Visual Recognition Branch}
\label{Visual}
The branch encodes visual features of the image according to the position self-enhanced embedding $P_{se}$, recognizes the scene text, and outputs the attention map ${AT}_m$ as visual cues. In this paper, ResNet50 is used as the backbone network to extract visual features. Given scene text image $I\in \mathbb{R}^{H\times W \times 3}$, the scene text recognition process of the visual recognition branch can be expressed as:
\begin{equation}
V_r=Res\left(I\right)\in \mathbb{R}^{\frac{H}{4}\times\frac{W}{4}\times C}\ \ \
\end{equation}
\begin{equation}
{AT}_m=softmax\left(\frac{P_{se}K_r^\mathsf{T}}{\sqrt C}\right)\in \mathbb{R}^{\frac{H}{4}\times\frac{W}{4}\times 1}
\end{equation}
\begin{equation}
F_v={AT}_m\ast V_r\ \ \
\end{equation}
where $C$ is the dimension of the feature channel, $Res$ stands for $Resnet50$, $K_r=U(V_r)$, and $U$ is $U-Net$. Based on the feature $F_v$, the recognition of the visual recognition branch can be formalized:
\begin{equation}
T_V=softmax(fully\left(F_v\right)) \in \mathbb{R}^{T\times cls}
\end{equation}
where $cls$ is the number of character classes, $fully$ is the fully connected layer and $softmax$ is activation function.

\subsection{Iterative Semantic Recognition Branch}
\label{Iterative}
\begin{figure}[ht]
  \centering
  \subfigure[Mult-Head self-mask attention]
  {\includegraphics[width=0.45\textwidth]{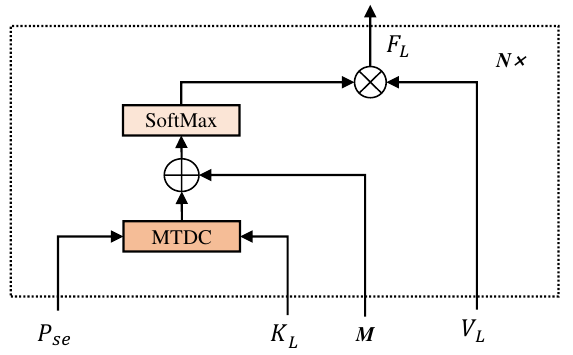}\label{fig:subfig3a}}   
  \subfigure[Mult-Head position enhanced self-mask attention]
  {\includegraphics[width=0.45\textwidth]{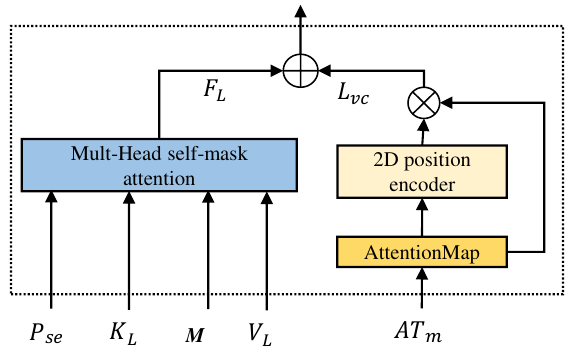}\label{fig:subfig3b}}  

  \caption{The structure of the Mult-Head self-mask attention \cite{9} and Mult-Head position enhanced self-mask attention. MTDC is short for Matrix multiplication, Transpose, Division, Channel square root. ${AT}_m$ comes from the visual recognition branch, and $L_{vc}$ is the representation of visual cues in the semantic space of the text.}  
  \label{figure3.}
\end{figure}

\noindent {\bfseries Language Recognition Module.} The language recognition module integrates visual cues to mine language information from the currently recognized text. The structure of the multi-head position enhanced self-mask attention module is shown in Fig. \ref{figure3.}. Inspired by \cite{9,6}, language models treat character reasoning as a fill-in-the-blank task. In order to prevent the leakage of its own information, the mask matrix $M\in R^{T\times T}$ is set. In this matrix, the elements on the main diagonal are negative infinity and the other elements are 0. Text semantic coding features can be obtained by position self-enhanced embedding $P_{se}$, visual recognition $T_v$ and attention map ${AT}_m$:

\begin{equation}
V_L=K_L=fully(T_V)\in \mathbb{R}^{T\times C}
\end{equation}
\begin{equation}
F_L=\ softmax\left(\frac{P_{se}K_L^\mathsf{T}}{\sqrt C}+M\right)V_L
\end{equation}
where $fully$ is the fully connected layer and $softmax$ is activation function. 
The language recognition can be expressed as:
\begin{equation}
L_{vc}={AT}_m\ast P_{s,c}^\prime \in \mathbb{R}^{T\times C}
\end{equation}
\begin{equation}
F_{vL}=F_L+L_{vc}
\end{equation}
\begin{equation}
T_L=softmax(fully\left(F_{vL}\right))\in \mathbb{R}^{T\times cls}
\end{equation}
where $P_{s,c}^\prime \in \mathbb{R}^{\frac{H}{4}\times\frac{W}{4}\times C}$ is the sine and cosine position code in two-dimensional space\cite{23,38}, and the sizes of its first two dimensions are consistent with the first two dimensions of ${AT}_m$. $T_L$ is visual text recognition results.

\begin{figure*}[t]
\centering
\includegraphics[width=1\textwidth]{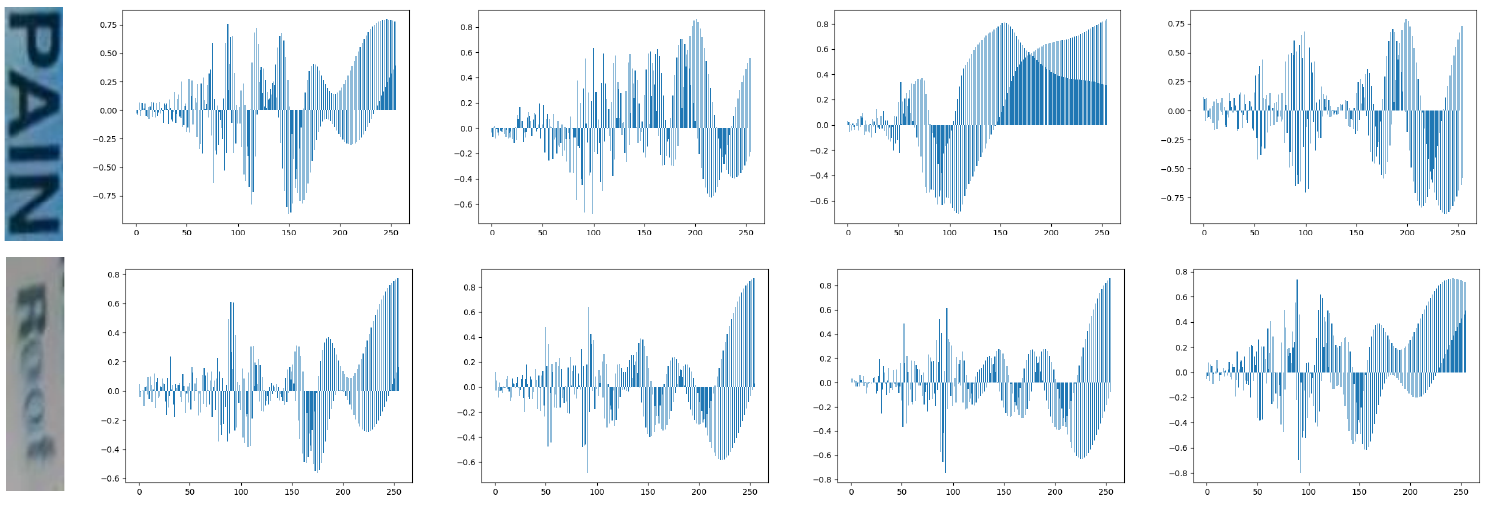}
\caption{Visualization of $L_{vc}$ for two scene text instances ("PAIN" and "Root"). The first figure in each row represents a scene text example, followed by the visual cues $L_{vc}$ representation corresponding to each character in the text. In the visualization diagram, the horizontal axis represents feature dimensions and the vertical axis represents the corresponding feature values.}
\label{figure4.}
\end{figure*}

$L_{vc}$ is considered as the representation of visual cues in the semantic space of the text, so can also refer to visual cues if not otherwise specified. The visualization of visual cues $L_{vc}$ for each character of two text instances "PAIN" and "Root" as shown in Fig. \ref{figure4.}. 

As can be seen from Fig. \ref{figure4.}, on the one hand, the visual cues $L_{vc}$ of different characters in the text have obvious differences. On the other hand, the $L_{vc}$ of the same character in a text tends to have similar expressions. However, due to the different positions, the expression also has a certain difference. This finding suggests that the use of visual cues can enhance the expression of text features.

\noindent {\bfseries Cross-Modal Fusion Gate.} In order to integrate the recognition of the visual recognition branch and language recognition module, this paper uses a simple fusion gate \cite{9,6}. The cross-modal fusion process can be expressed as:
\begin{equation}
F_g=\sigma([F_v,F_{vL}]Wg)
\end{equation}
\begin{equation}
T_F=F_v\ast\left(1-F_g\right)+F_{vL}\ast F_g
\end{equation}
where $W_g$ and $F_g$ are trainable superparameters and fusion weights, respectively.
\subsection{Training Objective Function}
\label{Loss}
In the training process, the recognition results of multiple modules need to be optimized, so we set a multi-objective loss function:
\begin{equation}
L=\gamma_vL_{Tv}+\frac{1}{N}\sum_{i=1}^{N}{{(\gamma}_l\times L}_{TL}^i+\gamma_f\times L_{TF}^i)
\end{equation}
where $L_{Tv}$ is the standard cross-entropy loss between the predicted probability  $T_V$ of the visual recognition branch and ground truth labels for the text. $L_{TL}^i$ and $L_{TF}^i$ are standard cross-entropy loss corresponding to the predicted probability $T_L$ of the language recognition module and $T_F$ of cross-modal fusion gate at the $ith$ iteration. $N$ is the number of iterations. $\gamma_v$, $\gamma_l$ and $\gamma_f$ are balanced factors. All of these balance factors are set to 1.0 in the experimental part of this paper.

\section{Experiments}
\subsection{Datasets}

\noindent {\bfseries Synthetic datasets.} MJSynth \cite{27} is a dataset generated by rendering text in scene images. It contained 9M texts, each of which is generated from a dictionary containing 90,000 English words, covering 1,400 Google fonts. SynthText \cite{28} dataset is originally proposed for the scene text detection task. In scene text recognition tasks, the dataset contains 8M images cut from detection task sets.

\noindent {\bfseries Regular datasets.} The three regular text datasets are ICDAR2013 (IC13) \cite{29}, Street View Text (SVT) \cite{30} and IIIT5k-words (IIIT) \cite{31}. IC13 consists of a training subset with 848 images and a validation subset with 1015 images. SVT comprises images sourced from Google Street View, which includes a training set of 257 samples and a validation set of 647 samples. IIIT includes a training subset of 2000 samples and a validation subset of 3000 samples. 

\noindent {\bfseries Irregular datasets.} Three irregular scene text datasets are ICDAR2015 (IC15) \cite{32}, SVT Perspective (SVTP) \cite{33} and CUTE80 (CUTE) \cite{34}. IC15 contains 4468 train samples and 2077 validation samples. SVTP is from Google Street View, which includes 238 images. 645 image blocks containing text are cropped from this dataset for scene text recognition tasks. The images in the CUTE dataset are collected from digital cameras or the Internet, with a total of 288 images, usually with high resolution and irregularly curved text. All the images are used to verify the algorithm.

\begin{table*}[t]
\caption{Text recognition accuracy comparison with other methods on six datasets. The current best performance on each dataset is shown in bold.}
\label{Table 2.}
\setlength{\tabcolsep}{1.4mm}{
\begin{center}
\begin{tabular}{l c|c c c|c c c | c}
\toprule
\multirow{2}{*}{Methods} & \multirow{2}{*}{Years}&\multicolumn{3}{c|}{Regular} &  \multicolumn{3}{c|}{Irregular} & \multirow{2}{*}{Params} \\ 
\cline{3-8}
\multirow{2}{*}&\multirow{2}{*}& IC13 & SVT & IIIT & IC15 & SVTP & CUTE \\ 
\midrule
MORAN \cite{14}  & 2019 & 92.4 & 88.3 & 91.2 & 68.8 & 76.1 & 77.4 & -\\
TRBA \cite{43} & 2019 & 92.3 & 87.5 & 87.9 & 77.6 & 79.2 & 74.0 & 49.6M \\
\hline

Textscanner \cite{22} & 2020 & 92.9 & 90.1 & 93.9 & 79.4 & 84.3 & 83.3 & 56.8M\\
RobustScanner \cite{6} & 2020 & 94.8 & 88.1 & 95.3 & 77.1 & 79.5 & 90.3 & -\\
SRN  \cite{2} & 2020 & 95.5 & 91.5 & 94.8 & 82.7 & 85.1 & 87.8 & 49.3 M\\
\hline

PIMNet \cite{21} & 2021 & 95.2 & 91.2 & 95.2 & 83.5 & 84.3 & 84.8 & -\\ 
VisionLAN \cite{20} & 2021 & 95.7 & 91.7 & 95.8 & 83.7 & 86.0 & 88.5 & 33M\\
ABINet \cite{9} & 2021 & 97.4 & 93.5 & 96.2 & 86.0 & 89.3 & 89.2 & 36.7M\\

\hline
SGBANet \cite{47} & 2022 & 95.1 & 89.1 & 95.4 & 78.4 & 83.1 & 88.2 & -\\
S-GTR \cite{45} & 2022 & 96.8 & 94.1 & 95.8 & 84.6 & 87.9 & 92.3 & 42.1M\\
ABINet-ConCLR \cite{40} & 2022 & 97.7 & {\bfseries 94.3} & 96.5 & 85.4 & 89.3 & 91.3 & -\\
\hline

CMFN  & Ours & {\bfseries 97.9} & { 94.0} &  {\bfseries 96.7} &  {\bfseries 87.1} & {\bfseries 90.1} & {\bfseries 92.0} & 37.5M\\
\bottomrule
\end{tabular}
\end{center}
}
\end{table*}

\subsection{Implementation Details}
In order to make the comparison with other SOTA algorithms \cite{9,40} as fair as possible, this paper adopts the experimental setup that is as close to these algorithms as possible. Before the scene image is input into the model, the size is adjusted to $32\times128$, and data enhancement pre-processings are adopted, such as random angle rotation, geometric transformation, color jitter, etc. In the experiment, the maximum length $T$ of the text is set to be 26. The text characters recognized are 37 classes, including 26 case-insensitive letters, 10 digits, and a token. The multi-head attention unit in the visual recognition branch is set to 1 layer. The language recognition module transformer is set as 4 layers and 8 heads. The experiment is implemented based on Pytorch. Two NVIDIA TITAN RTX graphics cards are used, each with 24GB of space. When MJSynth and SynthText are used for training, the training model provided by ABINet \cite{9} is used for initialization, the batch size is set to 200, the initial learning rate is ${1e}^{-4}$, and the optimizer is ADAM. The model is trained for a total of 10 epochs. The learning rate decays by one-tenth with each increase of epoch after 5 epochs.

\subsection{Comparisons with State-of-the-Arts}
In fact, due to the differences in training strategies and backbone, it is difficult to make an absolutely fair comparison between different methods. To be as fair as possible, only the results using the same or similar training strategies are analyzed and compared in this experiment. Specifically, each algorithm model is supervised learning, and the same datasets are used for training. 

The statistical results of our CMFN and other published state-of-the-art methods are shown in Tab. \ref{Table 2.}. Compared with ABINet-ConCLR \cite{40}, our CMFN improved 1.7\%, 0.8\%, and 0.7\% on the three irregular datasets IC15, SVTP, and CUTE, respectively. Compared to irregular datasets, the improvement on regular datasets is relatively small. This is because the accuracy of regular scene text recognition is already quite high (for example, the recognition accuracy of ABINet-ConCLR \cite{40} on the IC13 dataset has reached 97.7\%), leaving a relatively limited space for improvement. 

In conclusion, our CMFN outperforms state-of-the-art methods in recognizing irregular scene text and achieves comparable results for regular scene text recognition. 
 
\subsection{Ablation Study}
\begin{figure}[t]
\centering
\includegraphics[width=10cm,height=5cm]{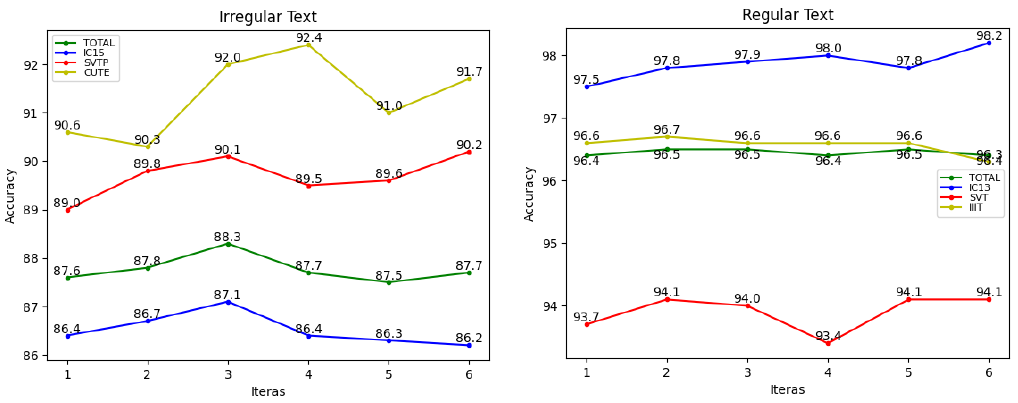}
\caption{ Text recognition accuracy of different iteration numbers. TOTAL indicates the statistic result of the three corresponding scene text datasets as a whole.}
\label{figure5.}
\end{figure}

\noindent {\bfseries Analysis of Iteration Number.} 
The visualization of text recognition results under different iterations number is shown in Fig. \ref{figure5.}. The overall recognition accuracy reaches 88.3\% on three irregular text datasets and 96.5\% on three regular datasets when the iteration number is set to 3. Further increasing the iteration number does not lead to significant improvement in accuracy. Therefore, the default iteration number for subsequent experiments is set to 3.

\noindent {\bfseries Analysis of the Position Self-enhanced Encoder and Visual Cues.} 
To verify the effectiveness of the position self-enhanced encoder, we compared its performance with learnable positional encoding \cite{48} and fixed positional encoding \cite{23}, as shown in Tab. \ref{Table1.}. The position self-enhanced encoder improved by 0.6\%, 1.3\%, and 1.0\% compared to fixed positional encoding, and by 1.0\%, 1.1\%, and 1.7\% compared to learnable positional encoding, on IC15, SVTP, and CUTE datasets. This indicates that our position self-enhanced encoder can bring higher recognition performance. Compared to the language module without visual cues, the inclusion of visual cues in the language module improves performance by 0.6\%, and 0.8\% on the IC15 and CUTE datasets. This demonstrates the effectiveness of the fusion of visual cues proposed in this paper.

\begin{table}[htbp]
\caption{Ablation study of position self-enhanced encoder. PSE, LPE, VPE are abbreviations for position self-enhanced encoder, learnable positional encoding and fixed positional encoding, respectively. TV, TL, TLn and TF represent the visual recognition branch, language recognition module with visual cues, language recognition module with visual cues exclude visual cues and the cross-modal fusion gate, respectively. }
\setlength{\tabcolsep}{8mm}{
\begin{center}
\begin{tabular}{l | c c c}
\hline
{Module} & {IC15} & {SVTP} & {CUTE}\\
\hline
VPE+TV+TLn+TF  & 85.9 &  88.7 &  89.2\\
VPE+TV+TL+TF &  86.5 &  88.8 &  91.0 \\
LPE+TV+TL+TF &  86.1 &  89.0 &  90.3 \\
PSE+TV+TL+TF & {\bfseries 87.1} & {\bfseries 90.1} & {\bfseries 92.0}\\
\bottomrule
\end{tabular}
\end{center}
}
\label{Table1.}
\end{table}

\begin{table}[htbp]
\caption{Ablation study of different modules. PSE represents the position self-enhanced encoder. TV(text of visual recognition), TL(text of language recognition) and TF(text of fusion gate recognition) represent the visual recognition branch, language recognition module, and the cross-modal fusion gate, respectively.}
\setlength{\tabcolsep}{8mm}{
\begin{center}
\begin{tabular}{l | c c c}
\hline
{Module} & {IC15} & {SVTP} & {CUTE}\\
\hline
PSE+TV  & 84.6 &  85.4 &  88.9\\
PSE+TV+TL &  84.3 &  89.8 &  89.9 \\
PSE+TV+TL+TF & {\bfseries 87.1} & {\bfseries 90.1} & {\bfseries 92.0}\\
\bottomrule
\end{tabular}
\end{center}
}
\label{Table 1.}
\end{table}

\begin{figure*}[t]
\centering
\includegraphics[width=1\textwidth]{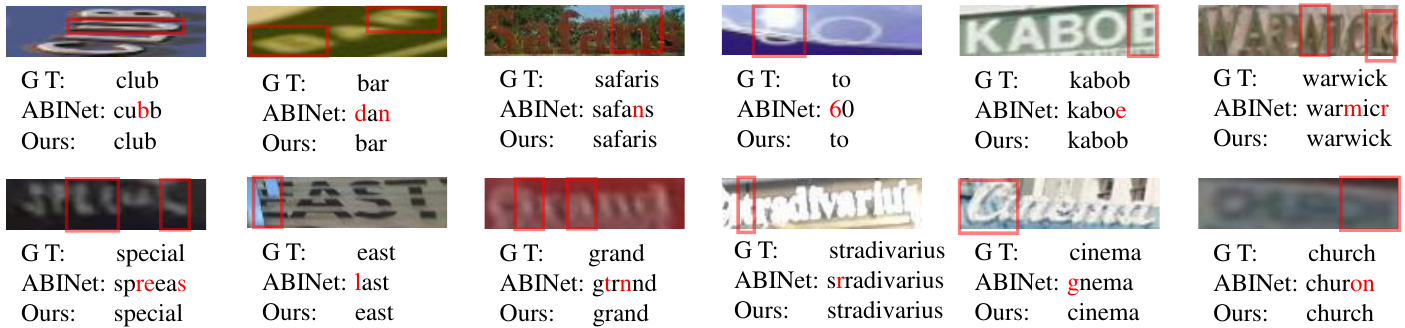}
\caption{Examples of successful recognition of irregular text. GT are the ground truth. ABINet and Ours are the recognition results corresponding algorithms, respectively.}
\label{figure6.}
\end{figure*}

\noindent {\bfseries Analysis of Different  Module.}
The ablation data of each module are shown in Tab. \ref{Table 1.}. The visual branch achieved 84.6\%, 85.4\%, and 88.9\% recognition accuracy on the IC5, SVTP, and CUTE. Compared to the visual recognition branch, the text recognition accuracy of the language recognition module decreased by 0.3\% on IC15 but increased by 4.4\% on SVTP and 1.0\% on CUTE. The cross-modal fusion gate achieved optimal recognition performance on IC15, SVTP, and CUTE, with improvements of 2.5\%, 4.7\%, 3.1\% compared to the visual recognition branch, and 2.8\%, 0.3\%, 2.1\% compared to the language recognition branch. This further shows that the modules in our model are valid.

\subsection{Qualitative Analysis}
For qualitative analysis, some recognition cases are visualized as shown in Fig. \ref{figure6.}. Compared with ABINet, CMFN has stronger recognition performance. For example, the visual features of "ri" in "safaris" are similar to those of "n", but "safans" is not a meaningful text, and our algorithm can correct "ri" to "n" according to the semantic relation, to correctly recognize the text. In summary, our CMFN has stronger text visual expression ability. For scene texts with insufficient visual features, the text also can be correctly recognized through semantic mining. Even for some scene texts that are difficult for human eyes to recognize, such as "church" and "grand", CMFN can also recognize them correctly.

\section{ Conclusion}

Inspired by human recognition of scene text, this paper proposes a cross-modal fusion network(CMFN) for irregular scene text recognition. The network mainly consists of three parts: a position self-enhanced encoder, a visual recognition branch, and an iterative semantic recognition branch. The position self-enhanced encoder encodes the position information of characters in the text. The visual recognition branch decodes the visual recognition text based on the visual features. The iterative semantic recognition branch simulates the way of integrating visual and semantic modes when a human recognizes scene text and improves the recognition performance of irregular texts by fusing visual features with semantic information. The experimental results show our CMFN not only achieves optimal performance in the recognition of irregular scene text but also has certain advantages in the recognition of regular scene text. In future research, we will explore how to design the recognition model based on knowledge reasoning.


%
%
%
\bibliographystyle{splncs04}

\end{document}